\documentclass[journal]{IEEEtran}
\usepackage{cite}
\usepackage{amsmath,amssymb,amsfonts}
\usepackage{algorithmic}
\usepackage{booktabs}
\usepackage{multirow}
\usepackage{array}
\usepackage{bigstrut}
\usepackage{graphicx}
\usepackage[outdir=./]{epstopdf}
\usepackage{textcomp}

\begin{document}
%
\title{Multiple Instance Segmentation in Brachial Plexus Ultrasound Image Using BPMSegNet }
%
%
%

\author{Yi Ding, \textit{Member, IEEE}, Qiqi Yang, Guozheng Wu, Jian Zhang, Zhiguang Qin, \textit{Member, IEEE}.
	\thanks{Corresponding authors: Yi Ding (yi.ding@uestc.edu.cn) and Jian Zhang (anesthesiology@zju.edu.cn).}
	\thanks{Yi Ding is with the School of Information and Software Engineering in University of Electronic Science and Technology of China, Chengdu, Sichuan, 610054, China; he is also with Institute of Electronic and Information Engineering of UESTC in Guangdong, Guangdong, 523808, China (e-mail: yi.ding@uestc.edu.cn). }
	\thanks{Qiqi Yang and Zhiguang Qin are  with the School of Information and Software Engineering in University of Electronic Science and Technology of China, Chengdu, Sichuan, 610054, China  (e-mail: qiqiyang@std.uestc.edu.cn; qinzg@uestc.edu.cn).}
	\thanks{Guozheng Wu is with National Natural Science Foundation of China, Beijing, China (e-mail: wugz@nsfc.gov.cn).}
	\thanks{Jian Zhang is with the Center of Anaesthesia surgery, Sichuan Provincial Hospital for Women and Children/Affilated Women and Children's Hospital of Chengdu Medical College, Chengdu, China (e-mail: anesthesiology@zju.edu.cn)}}
\maketitle

%
%

\markboth{Journal of \LaTeX\ Class Files,~Vol.~14, No.~8, August~2015}%
{Ding \MakeLowercase{\textit{et al.}}: Multiple Instance Segmentation in Brachial Plexus Ultrasound Image Using BPMSegNet }

\maketitle

\begin{abstract}
The identification of nerve is  difficult as structures of nerves are challenging  to image and to detect in ultrasound images. Nevertheless, the nerve identification in ultrasound images is a crucial step to improve performance of regional anesthesia. In this paper, a network called Brachial Plexus Multi-instance Segmentation Network (BPMSegNet) is proposed to identify different tissues (nerves, arteries, veins, muscles, etc,.) in ultrasound images. 
The BPMSegNet has three novel modules. The first is the spatial local contrast feature, which computes contrast features at different scales.  The second one is the self-attention gate, which reweighs the channels in feature maps by their importance. The third is the addition of a skip concatenation with transposed convolution within a feature pyramid network.
The proposed BPMSegNet is evaluated by conducting experiments on our constructed Ultrasound Brachial Plexus Dataset (UBPD). Quantitative experimental results show  the proposed network can segment multiple tissues from the ultrasound images with a good performance..
\end{abstract}

\begin{IEEEkeywords}
Brachial plexus segmentation, ultrasound image, deep learning, instance segmentation. 
\end{IEEEkeywords}

%
\IEEEpeerreviewmaketitle

\section{Introduction}

The ultrasound is one of the core diagnostic imaging  modalities and it is widely applied in diagnosis and treatment. For example, the ultrasound-guided injection of botulinum toxin A \cite{GarcUtilidad}, using ultrasound to diagnose neuralgic amyotrophy \cite{Marieke2018Ultrasound}, and peripheral nerve blockade (PNB) \cite{Barrington2013Ultrasound}. For PNB, the ultrasound  guidance has become the indispensable guidance modality \cite{Choi2014Evidence}. During the PNB, anesthesiologists perform regional anesthesia around the target nerve, where ultrasound images are used to locate nerve to promote pain management \cite{sites2006ultrasound}. Hence the accurate localization of nerve structures in ultrasound image is a critical step for effectively processing PNB procedures \cite{DENNY20051}. The clinicians, and in particular novices can benefit from assistance in interpreting these images.

Nowadays, deep learning algorithms, in particular convolutional networks, have rapidly become a methodology of choice for analyzing medical images  \cite{LITJENS201760, Shen2017Deep}. For example, \cite{ChenBrain,stackUnet,resnetBrain} design unique network architectures to segment the brain tumor in MRI images.  
For ultrasound image analysis,  \cite{MILLETARI201792} has proposed an approach for midbrain segmentation.  \cite{zhang2017image,zhao2017improved} use improved convolutional networks for nerve segmentation. 

However, there are challenges in nerve segmentation. First, the size of nerve is very small and  inconspicuous. Next, the noise disturbance in ultrasound imaging causes a reduction in image quality and degrades of  details like texture. Third, the low contrast with the neighbor also blurs the definition of boundaries of anatomical tissue.  
Besides the challenges in ultrasound imaging, identifying nerves in ultrasound image is also difficult for anesthesiologists. It requires anesthesiologists to fully understand the nerve and its surrounding anatomic landmarks (muscle, vein, and artery), and have extensive clinical experience \cite{Youngner2019}.

To solve these challenges, we have worked from two aspects.  
First, we  merge features from different scales to combine local and semantic information, and enhance the contrast information to highlight the small and inconspicuous objects.      
In addition, we have inspired by the clinical PNB operation. Anesthesiologists tend to identify the nerves by referring to its its surrounding anatomic landmarks  rather than locating the nerves directly. The identification of muscle, vein, and artery can help recognize the nerve in clinical practice. As a result, instead of  focusing  on  nerves, it is also important for network to learn the relationship among those tissues.

\begin{figure}[!t]
	\centerline{\includegraphics[width=\columnwidth]{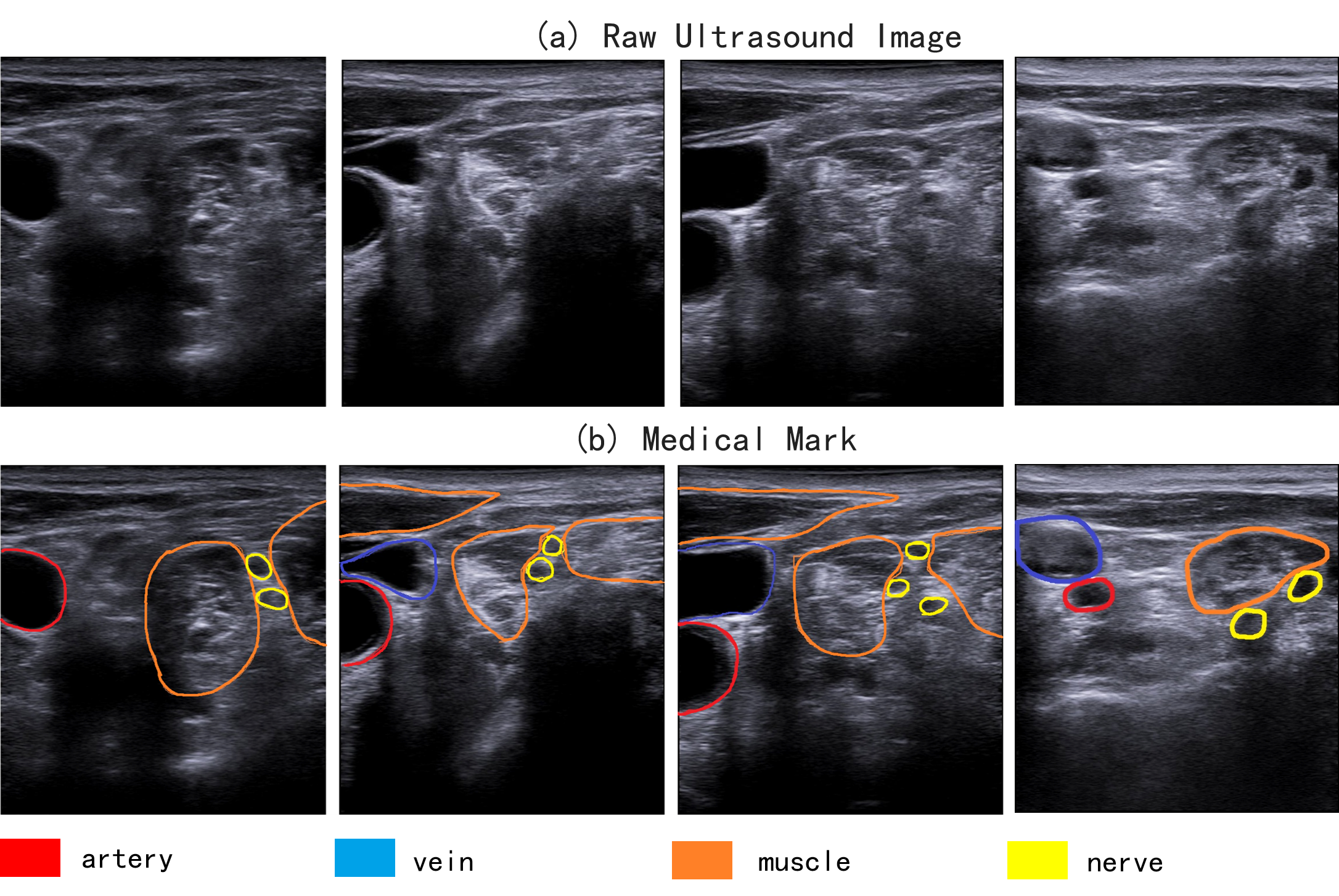}}
	\caption{
		(a) shows the raw ultrasound images collected  from different patients. (b) is the medical ground truth marked by anesthesiologist.
	}
	\label{UI}
\end{figure}	

In this paper, a Brachial Plexus Multi-instance Segmentation Network (BPMSegNet) is proposed based on the following insights. 
(1) Following the clinical procedure of anesthesiologists, the positional relationship  of nerve and its surrounding landmarks can help segmenting nerve. 
(2) The inconspicuous and small objects segmentation can be improved by multi-scale contrast features. 
The BPMSegNet is implemented by following the Mask R-CNN \cite{DBLP:journals/corr/HeGDG17} framework, besides, we propose some simple yet effective modules to improve the segmentation performance. 
First, we use Spatial Local Contrast Feature (SLCF) module to  aggregate  the contrast features in spatial and local resolution. 
Secondly, to select  channels   among feature maps and save useful information, the Self Attention Gate (SAG) is proposed.
Thirdly, the original cascade upsampling process in Feature Pyramid Network (FPN) \cite{DBLP:journals/corr/LinDGHHB16} is improved with transpose convolution and skip concatenation. 
Last, a ultrasound image dataset - Ultrasonic Brachial Plexus Dataset (UBPD) is constructed to evaluate the BPMSegNet, which contains nerve, muscle, vein, artery, and their corresponding  masks.

The contributions of this work are summarized as  follows: 

\begin{itemize}
	\item [1)] 
	A novel brachial plexus segmentation network, BPMSegNet, is developed to identify multiple instances for ultrasound image by applying  deep learning algorithm of instance segmentation and integrates with  prior medical knowledge in the clinical nerve identification process. 
	
	\item [2)] 
	We design   modules to improve  performance for nerve segmentation. The multi-scale contrast feature module combined with self-attention gate is developed to generate tailored features for ultrasound image. In addition, the upsampling mechanism in FPN is improved by adopting the skip connection to fine tune the information flow and increase the nonlinearity. Benefit from these designs, our method achieves higher nerve detection and localization accuracy than state-of-the-art methods.
	
	\item [3)] 
	We have built an ultrasound image dataset - UBPD.  This dataset dedicates to the segment brachial plexus and its surrounding anatomy. It consists 1055 ultrasound images with different targets and their corresponding labeled masks. 
\end{itemize}

The remainder of the paper is organized as follows. In Section \uppercase\expandafter{\romannumeral2} we briefly review the related work in instance segmentation and ultrasound image segmentation. In Section \uppercase\expandafter{\romannumeral3} we introduce the construction of our new dataset UBPD.  The mechanism of the proposed method is elaborated in Section \uppercase\expandafter{\romannumeral4}. The experimental results are presented in Section \uppercase\expandafter{\romannumeral5}.

\begin{figure*}[!t]
	\centerline{\includegraphics[width=0.8\textwidth]{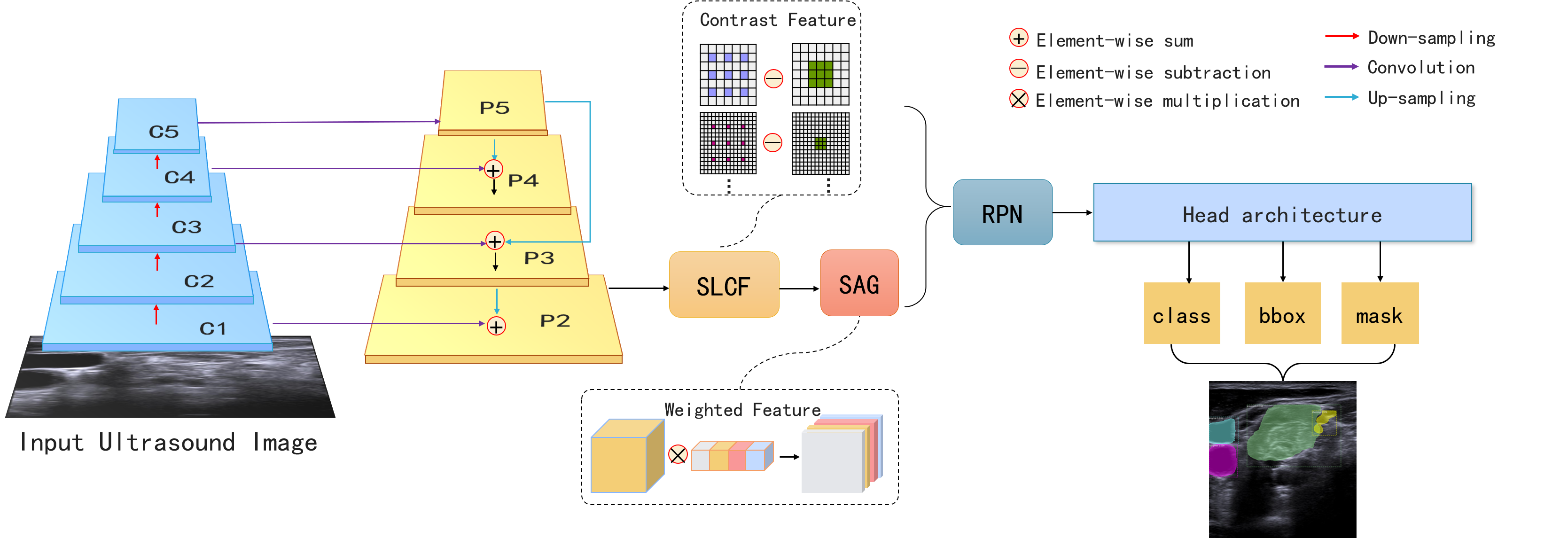}}
	\caption{The schematic of BPMSegNet. The Spatial Local Contrast Feature (SLCF) module learns contrast features  for ultrasound images, then Self Attention Gate (SAG) assigns adaptive weights to different channels. Last, the refined features are then utilized by region proposal network (RPN) and head architecture.}
	\label{fig_network}
\end{figure*}

\section{Related Work}

In this section, we briefly review the works in instance segmentation and  related improvements.

\subsection{Instance segmentation}
The idea of instance segmentation is to identify, localize and segment objects while distinguish different instances of the same category, which can be roughly divided into two main approaches: segmentation based methods and detection based methods. 

The detection based instance segmentation methods extended from object detection methods (e.g., Faster-RCNN \cite{ren2015faster}) to obtain detection instances, then add a mask branch to predict  the segmentation mask. In Mask RCNN, the network uses one additional branch to predict instance segmentation masks for the  detection instances generated from Faster-RCNN \cite{ren2015faster}. PANet \cite{liu2018path} proposed several improvements based on Mask R-CNN by adding bottom-up paths to facilitate feature propagation in. YOLACT \cite{DBLP:journals/corr/abs-1904-02689} achieves 29.8 mAP on Microsoft COCO dataset \cite{DBLP:journals/corr/LinMBHPRDZ14} by breaking instance segmentation into two parallel subtasks: generating a set of prototype masks and predicting per-instance mask coefficients. 

The segmentation  based method firstly  predicts the category labels of each pixel  and then groups them together to obtain pixel label of instance segmentation. \cite{liang2017proposal}  used  spectral clustering to cluster the pixels. \cite{zhang2015monocular} uses depth estimation and adds boundary detection  information  during  the clustering  procedure.

\subsection{Attention Mechanism}
Nowadays, the attention mechanism has  received extensive attention in both Natural Language Processing and computer vision. Non-local neural network \cite{wang2018non} first explored the research of attention mechanism in image processing, and give the intuitional instructions on the design of attention operations. 
Then the recently published DANet \cite{DBLP:journals/corr/abs-1809-02983}, OCNet \cite{DBLP:journals/corr/abs-1809-00916}, CCNet \cite{DBLP:journals/corr/abs-1811-11721}, PSANet \cite{zhao2018psanet}, and Local Relation Net \cite{DBLP:journals/corr/abs-1904-11491} explore the application of attention mechanism in semantic segmentation.

In attention mechanism, capturing long term dependencies \cite{hochreiter2001gradient} is of vital importance in deep neural networks. We can establish the connection between two pixels with a certain distance on the image (or on the feature map), which can be applied along space and time. But Convolution and recurrent neural network \cite{mikolov2010recurrent} often used in deep neural networks are operations performed on local areas, which are typical local operations. \cite{wang2018non} discuss the influences of non-local operations which can be designed to capture long-term dependencies.

The non-local operations consider the weighting of features between all locations when calculating the response at a certain position. So comparing to the general Euclidean distance, this dependency may reflect the relationship and connection between different positions, or the continuity of the same position in different dimensions. The design of non-local operation is summarized in Non-local Neural Networks as
\begin{equation}\mathit{y_{i}}=\frac{1}{C\left ( x \right )} \sum_{\forall j}f\left ( x_{i} ,x_{j}\right )g\left (x_{j}  \right ),\label{eq2}\end{equation}
where $i$ is the index of an output position whose response is to be computed. The $j$ is the index of other possible positions,  $x$ is the input (image,video, features) and $y$ is the computed output signal of position $i$. The function $f$ computes a scalar that represents the relationship between the position $i$ and all the others $j$. Lastly, function $g$ computes a representation of the input signal at the position $j$, and $C(x)$ is normalization factor.

\subsection{Ultrasound Images Segmentation}
The work \cite{noble2006ultrasound} has reviewed techniques developed for ultrasound image segmentation, including thresholding, level sets, active contours and other model-based methods. e.g. In \cite{cary2014brachial}, longitudinal and transverse view ultrasound image sequences of brachial arteries were segmented  by feed forward active contour (FFAC). 

For deep learning methods, many studies have shown the advantages of adopting the deep learning algorithm for ultrasound image segmentation, especially in accuracy and efficiency. In recent papers,  segmentation methods based on machine learning are proposed to achieve pixel-wise segmentation. There are various types  of neural networks designed and applied for ultrasound image segmentation. 
\cite{LEI2018178} uses boundary regularized convolutional encoder–decoder network to segment breast anatomy, \cite{noe2017high} and \cite{chauhan2016diagnostic} design the handcrafted features for ultrasound image, and \cite{carneiro2011segmentation,milletari2017hough} use deep learning based method to meet different clinical needs. The segmentation in 3D ultrasound image has also been studied in recent work \cite{looney2017automatic,yang20163d}.  

For specific deep learning solutions in brachial plexus segmentation, there are studies which use semantic segmentation methods. For example, \cite{zhang2017image, zhao2017improved, DBLP:journals/corr/HafianeVD17, Baby2017Automatic} propose their network to segment nerve in ultrasound images. \cite{zhang2017image} combines the features from shallow and deep layers through multi-path information confusion, and use dilated convolution to enlarge the receptive field in deep layer.   
\cite{zhao2017improved} proposes a U-net like network too, and takes advantage of inception modules and batch normalization instead of ordinary convolutional layers.
There are few limitations in these works. First, these studies transfer the commonly used neural network structures in natural image to ultrasound image segmentation, and do not consider the low resolution and low contrast of ultrasound image. 
Moreover, in clinical practice, the brachial plexus identification is highly related to the identification of its surroundings. They only focus on the nerve, and do not take the important relevance between nerve and other tissues into account.
In clinical practice, the brachial plexus identification in ultrasound image is very important for visual navigation. Hence, we construct the dataset for brachial plexus and its surrounding anatomical structure, and explore the application of deep learning on brachial plexus segmentation to assist anesthesiologist in PNB process.

\section{Dataset}
We have constructed an Ultrasonic Brachial Plexus Dataset (UBPD) that focuses on segmenting brachial plexus and its surrounding anatomical tissues. 

The dataset is collected  from 101 patients  by professional anesthesiologists. 
Firstly, there are two ultrasound devices (SIEMENS ACUSON NX3 Elite and Philips EPIQ5) used for data collection,  and anesthesiologists use high frequency probe  to scan the targets. The acquisition method is the same as  clinical practice. 
The anesthesiologists initially place probes from the middle of the right neck to find suitable view, and capturing depth is 4cm. When the video shows the internal jugular vein, common carotid artery, anterior oblique muscle and brachial plexus coexisting, the anesthesiologist will slowly slide the probe down. 
The ultrasound video is recorded for a total of 8 seconds. By following this method, we collected a total of 101 ultrasound videos.
Next, the ultrasound images are captured by extracting frames in these videos. We randomly extract 10 to 15 frames  from each ultrasound video, and the same time we ensure there is at least one target in the selected frame. 
Together we have obtained an ultrasound dataset with a total of 1052 images. 

Each ultrasound image contains targets and their corresponding annotated masks. The Labelme \cite{russell2008labelme} is adopted for manual labeling, and the ground truth labels are marked by the anesthesiologists. 
Part of the masks and images in UBPD are shown in Figure \ref{UI}, there are 4 categories including nerve, muscle, vein, and artery. These tissues varies in shape and size, and have different characteristics. 
The vein and artery are salient in ultrasound imaging,  and their edges are clear and large in size. On the contrary, the size of nerve is small, and it is inconspicuous.

During the labeling process, anesthesiologist has concluded that the nerve has the following characteristics:  bright edges, dark in internal, its diameter is within 3mm, and it is continuous beaded or honeycomb. In clinical practice, the anesthesiologist will first identify the arteries and veins, then identify the muscle, and finally the nerve. The target recognition process of the anesthesiologists is consistent with their examination procedures in clinical anesthesia. 

In our experiments, the dataset is divided into training  and testing dataset. The training dataset consists of 955 images  collected from 91 patients, and the testing dataset consists of 97 images  collected from 10 patients.

\section{The BPMSegNet}


In this paper, the Brachial Plexus Multi-instance Segmentation Network (BPMSegNet) is proposed for brachial plexus segmentation.
The shape of the nerve shown in ultrasound image is highly variable, and nerves with clear edges and brightness contrast are easier to be recognized. This observation suggests that the successful identification of nerves needs to highlight and enhance the contrast information. 
In addition, in clinical nerve identification process, the anesthesiologists rely on the position of its surroundings to locate the suitable view for nerves.  
This indicates that considering positional relationship among these anatomical targets can help the nerve identification.
Inspired by the above observations, we design  some components for the proposed network. 
First, the SLCF module is proposed to save the contrast information in spatial and local resolution for multi-scale features aggregation. 
Secondly, to further select useful information among a large number of channels produced by SLCF, the SAG is proposed to filter redundant information. 
The detailed description of the proposed framework can be found in the following subsections.

\subsection{The Structure of BPMSegNet }
The structure of BPMSegNet  is illustrated in Figure \ref{fig_network}. It takes a single image  as input and extract multi-level features (C2, C3, C4, C5) by the feature extraction network. 
In this paper,  the \textit{output stride} is expressed as the ratio of image resolution to the feature map resolution.  For example ,the size of C5 is 32 times smaller  than the input (because of 5 downsampling operations), so  the \textit{output stride} is 32. 
Although C5 has the lowest resolution, its semantic information is the most abundant.
While feature map C2 is of low-level, which includes local  and detailed information. 
Then, the BPMSegNet uses a U-shape structure  with skipping  transposed convolution that akin to FPN for fusing multi-scale features. This upsampling branch outputs feature maps with \textit{output stride} of 32, 16, 8, and 4 respectively, which are denoted as the P5, P4, P3, and P2. 

Then, the feature map in P2  is fed into SLCF module to learn multi-scale contrast features. The output feature maps of SLCF are further selected and filtered by the SAG. Next these feature maps are inputed into region proposal network (RPN) \cite{ren2015faster} to scan the image and locate objects. The RPN candidate bounding boxes that are more likely to contain foreground objects, and then performs multi-classification and bounding refinement on candidate bounding boxes by non-maximum suppression (NMS) \cite{girshick2015deformable}. Lastly, the FCN network \cite{long2015fully} in the mask branch classifies the segmentation mask and generates boundary boxes.

\begin{figure*}[!t]
	\centerline{\includegraphics[width=0.9\textwidth]{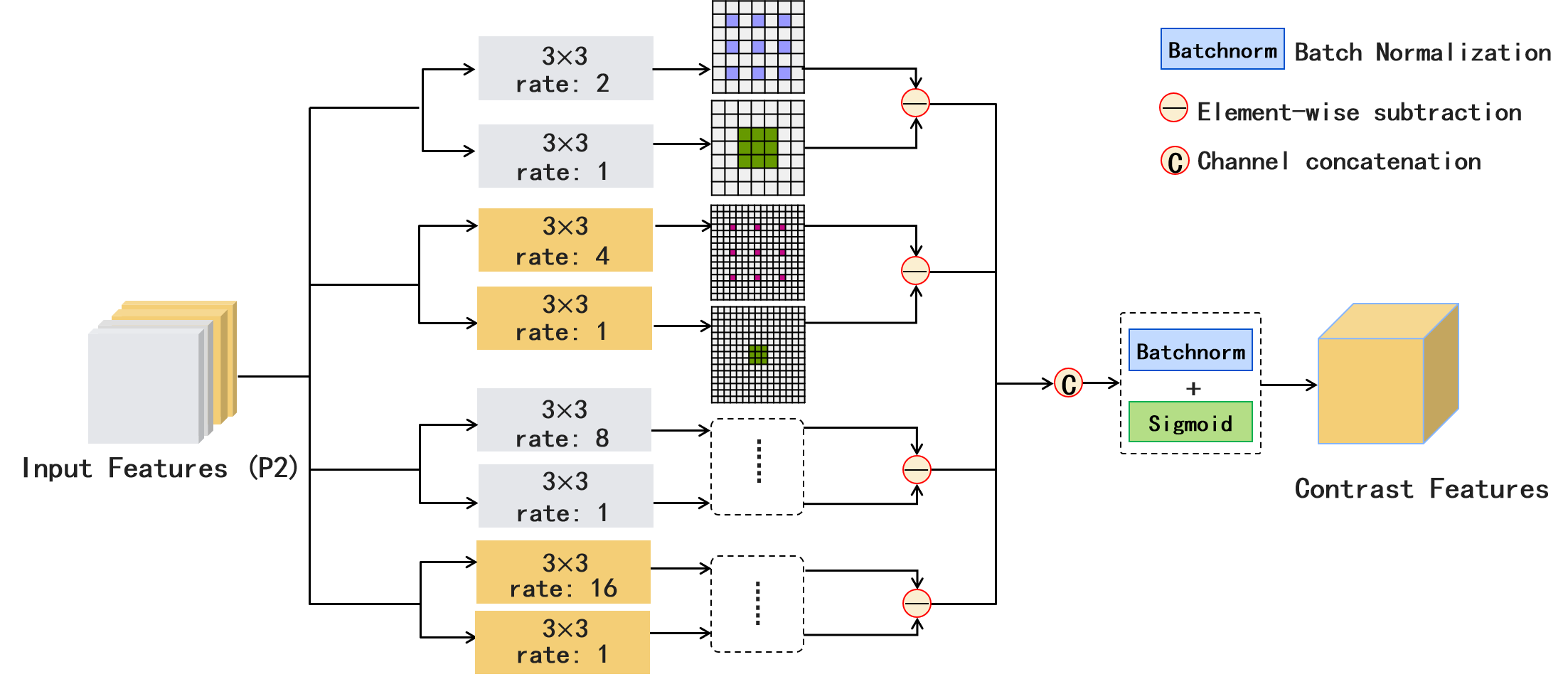}}
	\caption{The Spatial Local Contrast Feature (SLCF) module.}
	\label{fig1}
\end{figure*}

\subsection{Spatial Local Contrast Feature}
We have observed that human eyes are sensitive to the contrast (brightness, shape) and adaptively adjust the size of pupils to perceive different visual information. It is easier for us to identify items with clear edges and with high contrast, and this makes the concept of contrast very essential for computer vision and image recognition.

There are two types of contrast features adopted in the segmentation task: spatial contrast and local contrast feature. The spatial contrast refers to the contrast between large regions, and it reveals the global and semantic information of the image. The local contrast refers to the contrast between smaller regions.
For salient objects (e.g. artery and vein in Figure \ref{UI} (b)), they have clear edges, large shapes and other histology features. In addition, they are more sensitive to the semantic information and more advantage in segmentation task \cite{li2016deep}. However, the advantages of salient objects result in the challenge for segmenting the inconspicuous and small objects. 
When aggregating features of different scales using upsampling and element-wise sum, the features of pixels in inconspicuous objects will be dominated by the features of salient objects. Some information of inconspicuous objects would be ignored in the final prediction, resulting in incorrect labeling for pixels at certain locations. 
For nerve segmentation task, the size of  objects is across a large range of scales, and it is important to consider both spatial and local contrast features. Therefore, the Spatial Local Contrast Feature (SLCF) is proposed to effectively aggregate spatial and local contrast features.

The structure of proposed SLCF module is shown in Figure \ref{fig1}. Given the input feature maps, the SLCF aggregates the contrast features at four different scales by four convolutional branches (with dilate rate of 2, 4, 8, 16). 
The SLCF is only used on P2, because the size of P3, P4, and P5 is relatively smaller, it will produce a lot of useless information when applying convolutional (Conv) operations with a large kernel size.  
P2 has the largest resolution and the most detailed features, which is desirable for generating contrast features.

There are four parallel contrasting blocks in the SLCF module. As shown in each branch, contrast features are generated by subtracting feature maps generated by convolution with different perception fields, which is expressed as follows 

\begin{equation}\mathit{SLCF} = (f \ast _{r}k)\left ( p \right )- (f \ast k)\left ( p \right ).\label{eq1}\end{equation}
The dilated convolution $\ast _{r}$  between input  $f$ and kernel $k$ and dilation factor $r$ is defined as
\begin{equation}(f \ast _{r}k)\left ( p \right )=\sum _{s+rt=p}f(s)k(t).\label{eq1.1}\end{equation}
The normal convolution $\ast$ is defined as
\begin{equation}(f \ast k)\left ( p \right )=\sum _{s+t=p}f(s)k(t).\label{eq1.2}\end{equation}
Here $f$ is the input feature map, and  $k$ represents the kernel (kernel size $3\times3$).  
The operator $\ast _{r}$ refers to $r$-dilated convolution, and the operator $\ast$ is the plain 1-dilated convolution.  
The subtraction results $\mathit{SLCF}$ are the expected contrast features, and then multi-scale contrast features produced by these four blocks will be concatenated together.

In the SLCF module, the  contrast features generated by different convolutions have four different scales. The convolution with a  larger dilation rate corresponds to a larger receptive field, and the size of the receptive field reflects the  abstraction degree of features. As shown in Figure \ref{fig1}, in these four parallel contrasting blocks the dilation rate $r$ is set to 2, 4, 8, and 16 respectively, to obtain feature maps with a long-range of receptive fields. When the dilation rate is small, the SLCF can contrast the local features with more discriminative and detail information. When the dilation rate is large, the SLCF are making contrast with the spatial features, which contain more global and semantic information. By fusing the outputs of four parallel contrasting blocks, the SLCF can generate more tailored contrast features for multi-scale objects, and these features can restrain the network to generate customized details both for  salient and inconspicuous objects. Finally, these multi-scale contrasting feature maps are concatenated together, and then be refined by SAG.

\begin{figure}[!t]
	\centerline{\includegraphics[width=\columnwidth]{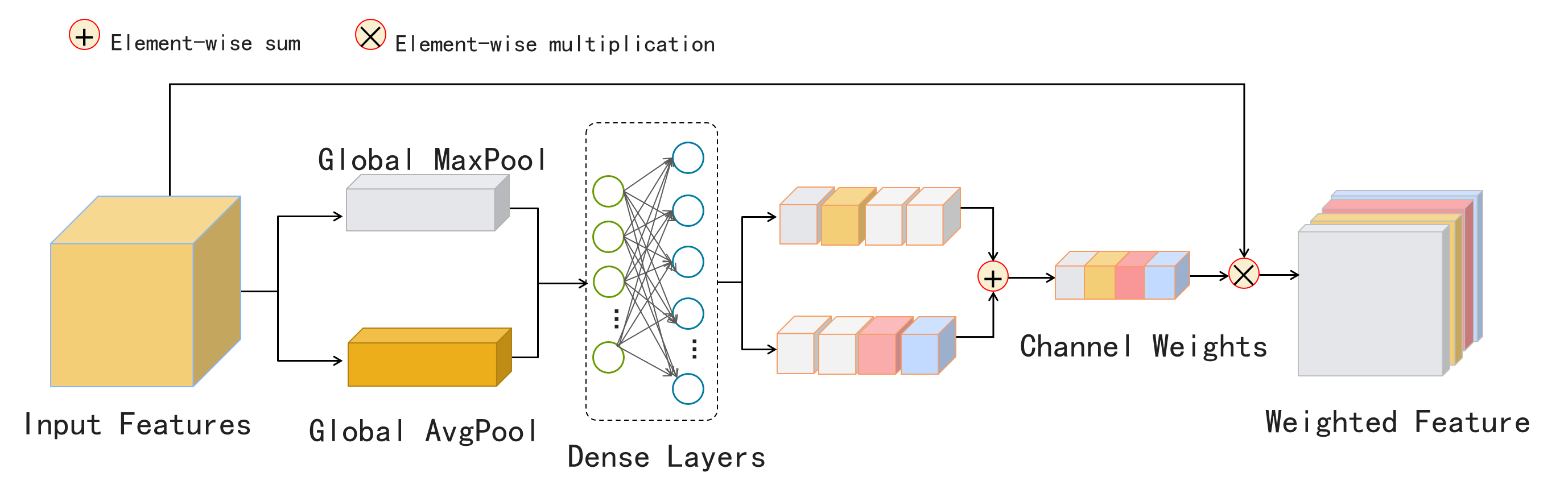}}
	\caption{ The structure of SAG module, in which the features are combined linearly with adaptive weights. 
	}
	\label{fig2}
\end{figure}

\subsection{Self Attention Gate}
The output of SLCF is a feature map matrix with 512 channels. However, too many channels will cause the parameters of network to sharply increase, and result in a large amount of computation consumption. In addition, over-trained parameters can lead to the recession of training effectiveness and even the performance degradation. In this case, obtaining more useful and streamlined information from feature maps can improve the performance of network and reduce computational overhead.

To address this problem, the Self Attention Gate (SAG) is proposed to establish the connections with positions in different channels and highlight those useful features. The SAG can  capture the relationships of different positions in different channels by adopting the non-local design mechanism. Comparing to continuous stacking convolution operations, the SAG can compute the long-term dependencies \cite{vaswani2017attention} (the relationship between the channels) rather than the relationship of pixels in one channel. 

In fact, such a method of establishing a connection between channels can be regarded as a self-attention \cite{zhang2018self} mechanism. The self-attention mechanism calculates the aggregates of attention scores at certain positions by interacting within the feature maps (here refers to the channels). It pays attention to all positions and obtains their weighted average in a high-dimensional embedding space.  Different channels can be considered as a high-dimensional space that expresses features in more abstract and diverse way.  The SAG follows the design of self-attention, which is achieved by focusing on all positions in a feature map and taking their weighted average on the channel dimension. 

The detail structure of SAG is shown in Figure \ref{fig2}, and the output of SAG (${\mathbb{F}}'$) is defined as

\begin{equation}{\mathbb{F}}'=W\left ( \mathbb{F} \right )\otimes \mathbb{F},\label{eq6}\end{equation}
\begin{equation}{W_{i}}\left ( \mathbb{F} \right )=\sigma \left ( \Theta\left ( F_{i}^{GAP} \right )+ \Theta\left ( F_{i}^{GMP} \right )\right ),\label{eq3}\end{equation}
where $\mathbb{F}$ is the input feature maps, $W\left ( \mathbb{F} \right )$ represents the updated weights of channels in input $\mathbb{F}$, and ${W_{i}}\left ( \mathbb{F} \right )$ is the weight for the $i$-th channel. $\Theta$ represents the parameters of dense layers, and $\sigma$ is sigmoid activation. 
The final output ${\mathbb{F}}'$ is the dot product of input $\mathbb{F}$ and $W\left ( \mathbb{F} \right )$. 

Firstly, the $i$-th channel is  compressed into two vectors by Global Average Pooling (GAP) and Global Max Pooling (GMP) respectively, which contain  global and mean information. The GAP and GMP are defined as
\begin{equation}\mathit{F^{GAP}}=\frac{\sum_{m, n}^{k} x_{m, n} }{n},\label{eq4}\end{equation}
\begin{equation}\mathit{F^{GMP}}=max(x_{m, n}).\label{eq5}\end{equation}
Here $(m, n)$ is the pixel index in feature map (size $k$ $\times$ $k$), and $x_{m, n}$ stands for the value of location $(m, n)$. 

Then the two vectors are input to a simple neural network to update the  separately. 
This neural network consists of two dense connected layers, and each layer has different number of  neural units. 
There are 64 units in the first layer, and 512 units in the second layer. This network can reassign  the weights to $F^{GAP}$ and $F^{GMP}$  by update its parameters.
The SAG increases the weight of effective channels and reduces the weight of invalid ones, so that the network can  choose  useful channels from feature maps. The output of SAG is with the same size as the input, so the SAG can be applied to multiple layers and easily embedded into other network architectures.  
More importantly, the SAG can discover the internal relationship among pixels in different channels without increasing many parameters.

\begin{figure*}[!t]
	\centerline{\includegraphics[width=0.9\textwidth]{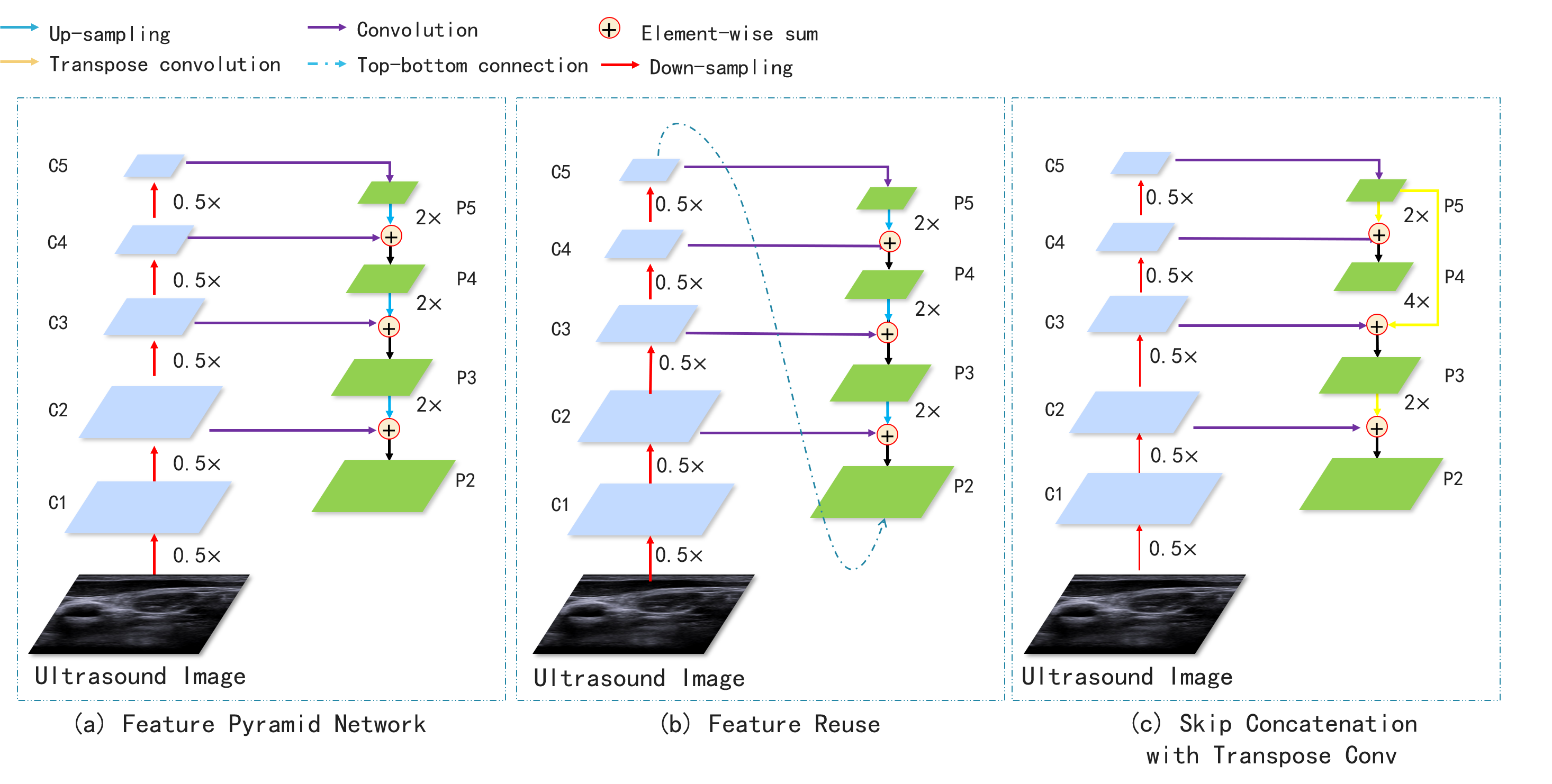}}
	\caption{
		(a) is the Feature Pyramid Network (FPN) where the feature maps are gradually upsampled by bilinear interpolation with a factor of 2. (b) enhances the feature reuse in FPN. (c) is our improved upsampling structure, and the feature maps are upsampled by transpose convolution, and the p5 is directly connected to the p2  with factor of 4. 	
	}
	\label{fig3}
\end{figure*}

\subsection{Skip Concatenation with Transpose Convolution }
For the feature maps, different layers compute features maps at different levels and different scales. A larger \textit{output stride} indicates  that it has low-resolution, and high semantic information.
Conversely, the feature map with smaller \textit{output stride} (e.g. P2) has high-resolution and finer information, and they are important for restoring the edge and boundary of objects.
Although all the feature maps  are at different scales and have huge semantic gaps generated by different layers in the network. 
The high-resolution maps have low-level features can damage their representational capacity for object recognition, and  we need to combine both high-resolution and low-resolution features to do a robust segmentation/detection at different scales. Therefore a skip concatenation with transpose convolution (SC) is proposed to combine low-resolution feature maps with high-level semantic information.

As shown in Figure \ref{fig3} (a), in the FPN, the low-resolution feature  maps  are  upsampled and then connected with larger resolution. The feature maps are upsampled by bilinear interpolation, which can be consider as a  linear decoder module. 
Base on the FPN, Figure \ref{fig3} (b) enhances the feature reuse by adding another connection between the lowest-resolution and highest-resolution feature maps. 

The proposed improved method is illustrated in \ref{fig3} (c). We change the connection between P4 to P3, and re-connect the P5 and P3.  The feature map in P5 has the highest level of semantic information and globality.  There is a huge semantic gap between P5 and P2, and connecting P5 directly to P3 can propagate high-level semantic information while prevent the information being discarded by many upsampling operations. 
Besides, instead of using bilinear interpolation, the transpose convolution is adopted to increase nonlinearity.  
The size of transpose convolution output is: s$\times$(i-1)-2p+k.
the $s$ is stride of upsampling, $i$ is size of input, $p$ is padding size, and $k$ is size of kernel.  The $p$ is set to 0, $k$ and $s$ is set to 2, except the skip concatenation from P5 to P3, in which $s$ and $k$ is set to 4.
By combining bottom-up pathway with top-down pathway using lateral connections, the high semantic information in the low-resolution feature maps are connected to high-resolution feature maps, and the  skip concatenation saves the high-level information while keep a balanced budget between efficiency and effectiveness.

\section{Experiment}

This section presents the performance of  the proposed method through different experiments. These experiments are all performed on the UBPD. The UBPD is divided into a  training dataset (consists of 955 ultrasound images) and a test set (consist of 97 ultrasound images), which contains 4 categories and the background. In order to evaluate the performance of  proposed  network, extensive experiments are conducted  on the test set in both  detection and  segmentation task.

\subsection{Implementation Details}
\textbf{Setup.} The proposed network is implemented with Tensorflow backend and runs on Nvidia GTX-1080Ti. The backbones used in the feature extraction network are ResNet101, ResNet50, and VGG-19 respectively. In addition, to constrain the parameters of the whole network, the number of channels for each block is reduced to half the number as that used in the normal network, thereby greatly reducing the overall parameters of the network. 

\textbf{Training.} The network is trained end-to-end on the training dataset of UBPD with 2 images per GPU, the SGD optimizer with fixed momentum 0.9 and weight decay 0.0001 is adopted for the implementation. The learning rate is initialized to 0.01, and decays by a factor of 10 for each 10th epoch. 
Since the ultrasound images are collected from different ultrasonic devices and the sizes of the images are varied, in the training process, the images are firstly resized into  640$\times$640 by using bilinear interpolation.
The training dataset is relatively smaller than the natural image dataset, so the network is trained with a total of 100 epochs to avoid overfitting. 
Each module is not specifically pre-trained on any dataset or initialized. 
The feature extraction network consists of five blocks. Each block contains a series of convolution and downsample operations, which will reduce the size of the feature map into half  of its original size. The multi-scale outputs of the feature extraction network are saved and represented by C1, C2, C3, C4, and C5, respectively.

\subsection{Experimental Evaluation}
In this work, the performance of the proposed network is evaluated from three aspects: object detection,  segmentation, and  the computation efficiency.  For detection task, the goal is to  predict the bounding boxes (B-box) of  each  object  of  that  class  in  a  test  image.  For segmentation task, in each test image, predict the class of each pixel, or background if the object does not belong to one of the 4 specified categories.  

To measure the detection precision and segmentation precision, we use average precision (AP) (i.e the area under the precision/recall curve) over all categories as the metric.
The mask AP (AP$^{mask}$) is the metric for the segmentation, and the B-box  AP (AP$^{B-box}$) is the metric of detection. To compute this metric, a group of  Intersection over Union (IoU) thresholds (0.5, 0.6, and 0.7) are used. The IoU is the overlap ratio between  prediction and ground truth. If the overlap ratio is lower than the set IoU threshold, the prediction is considered as false. The IoU overlap is defined as follow:

\begin{equation}IoU= \frac{I_{g} \cap  I_{p}}{I_{g} \cup  I_{p}},\label{eqIou}\end{equation}
where $I_{g}$ is the area of ground truth mask (ground truth B-box), and $I_{p}$ denotes the area of predicted mask (predicted B-box).

As shown in Table \ref{backbone} \ref{SEGMENTATION} \ref{DETECTION} \ref{others}, there are AP$_{50}$, AP$_{60}$, AP$_{70}$, and AP. Specifically, AP$_{50}$, AP$_{60}$, AP$_{70}$ represent  the AP computed at IoU threshold  0.5, 0.6, and 0.7 respectively, and AP in the last column means the average AP over IoU threshold = 0.50 : 0.05 : 0.95.

\subsection{Ablation Study }
To verify the validity of the proposed module, the ablation studies are implemented by removal of one  or more modules. 
The standard metrics consist of AP, AP$_{50}$, AP$_{60}$, and AP$_{70}$ for both B-box and segmentation masks.  To  reveal the different performance of each proposed component, except for the differences specified in each ablation experiment, the other settings are remained consistent (following the training method in Implementation Details). The  results of different  studies are shown in Table \ref{SEGMENTATION} and Table \ref{DETECTION}. 
In the following, the performance of each proposed  module is discussed in detail.

\begin{table}[htbp]
	\centering
	\caption{The comparison of effects of different feature extraction network on Detection AP (AP$^{B-box}$) and Mask AP (AP$^{mask}$).}
	\renewcommand{\arraystretch}{1.1} 
	\begin{tabular}{c|c|c|c|c}
		\toprule
		\multicolumn{5}{c}{(a) Mask AP of different feature extraction network} \\
		\midrule
		\multirow{2}[4]{*}{Conditions } & \multicolumn{4}{c}{AP$^{mask}$} \\
		\cmidrule{2-5}          & AP$_{50}$  & AP$_{60}$  & AP$_{70}$  & AP  \\
		\midrule
		ResNet-50 backbone &  43.81 &  39.22 &  31.67 &  28.31 \\
		\multicolumn{1}{p{9em}|}{ResNet-101 backbone} &  45.47 &  39.45 &  31.94 &  28 \\
		VGG-19 backbone &  49.33 &  45.19 &  35.21 &  30.97 \\
		\toprule
		\multicolumn{5}{c}{(b) Detection AP of different feature extraction network} \\
		\midrule
		\multirow{2}[4]{*}{Conditions } & \multicolumn{4}{c}{AP$^{B-box}$} \\
		\cmidrule{2-5}          & AP$_{50}$  & AP$_{60}$  & AP$_{70}$  & AP  \\
		\midrule
		ResNet-50 backbone &  50.77 &  45.6 &  38.63 &  32.58 \\
		\multicolumn{1}{p{9em}|}{ResNet-101 backbone } &  58  &  48.71 &  38.4 &  33.11 \\
		VGG-19 backbone &  59.09 &  51.45 &  34.57 &  33.54 \\
		\bottomrule
	\end{tabular}%
	\label{backbone}%
\end{table}%

\begin{table}[htbp]
	\centering
	\caption{The segmentation results of ablation experiments evaluated on UBPD dataset. The best results are
		highlighted in bold.}
	\begin{tabular}{c|c|c|c|c}
		\toprule
		\multicolumn{5}{c}{(a) SLCF performance} \\
		\midrule
		\multirow{2}[4]{*}{Conditions } & \multicolumn{4}{c}{AP$^{mask}$} \\
		\cmidrule{2-5}          & AP$_{50}$  & AP$_{60}$  & AP$_{70}$  & AP  \\
		\midrule
		\multicolumn{1}{c|}{ResNet-101 backbone} &  45.47 &  39.45 &  31.94 &  28 \\
		VGG-19 backbone &  49.33 &  45.19 &  35.21 &  30.97 \\
		\midrule
		\multicolumn{1}{c|}{ResNet-101+SLCF} &  49.45 &  41.06 &  33.75 &  30.43 \\
		VGG-19+SLCF &  54.53 &  46.27 &  36.15 &  33.21 \\
		\bottomrule
		
		\toprule
		\multicolumn{5}{c}{(b) SAG performance} \\
		\midrule
		\multirow{2}[4]{*}{Conditions } & \multicolumn{4}{c}{AP$^{mask}$} \\
		\cmidrule{2-5}          & AP$_{50}$  & AP$_{60}$  & AP$_{70}$  & AP  \\
		\midrule
		\multicolumn{1}{c|}{ResNet-101 backbone} &  45.47 &  39.45 &  31.94 &  28 \\
		VGG-19 backbone &  49.33 &  45.19 &  35.21 &  30.97 \\
		\midrule
		\multicolumn{1}{c|}{ResNet-101+SAG} &  51.1 &  44.13 &  34.44 &  31.8 \\
		VGG-19+SAG &  52.05 &  46.34 &  36.07 &  33.9 \\
		\bottomrule
		
		\toprule
		\multicolumn{5}{c}{(c) SLCF+SAG performance} \\
		\midrule
		\multirow{2}[4]{*}{Conditions } & \multicolumn{4}{c}{AP$^{mask}$} \\
		\cmidrule{2-5}          & AP$_{50}$  & AP$_{60}$  & AP$_{70}$  & AP  \\
		\midrule
		\multicolumn{1}{c|}{ResNet-101 backbone} &  45.47 &  39.45 &  31.94 &  28 \\
		VGG-19 backbone &  49.33 &  45.19 &  35.21 &  30.97 \\
		\midrule
		\multicolumn{1}{c|}{ResNet-101+SLCF+SAG} &  51.1 &  44.13 &  34.44 &  31.8 \\
		VGG-19+SLCF+SAG &  54.06 &  48.62 &  36.55 &  35.2 \\
		\bottomrule
		
		\toprule
		\multicolumn{5}{c}{(d) SLCF+SAG+SC performance} \\
		\midrule
		\multicolumn{1}{c|}{\multirow{2}[4]{*}{Conditions }} & \multicolumn{4}{c}{AP$^{mask}$} \\
		\cmidrule{2-5}    \multicolumn{1}{c|}{} & AP$_{50}$  & AP$_{60}$  & AP$_{70}$  & AP  \\
		\midrule
		ResNet-101 baseline &  45.47 &  39.45 &  31.94 &  28 \\
		\multicolumn{1}{c|}{VGG-19 backbone} &  49.33 &  45.19 &  35.21 &  30.97 \\
		\midrule
		ResNet-101+SLCF+SAG+SC &  50.55 &  46.19 &  39.53 &  33.56 \\
		VGG-19+SLCF+SAG+SC & \textbf{ 58.02} & \textbf{ 48.92} & \textbf{ 42.06} & \textbf{ 37.18} \\
		\bottomrule
		
	\end{tabular}%
	\label{SEGMENTATION}%
\end{table}%

\begin{table}[htbp]
	\centering
	\caption{The detection results of ablation experiments evaluated on UBPD dataset. The best results are
		highlighted in bold.}
	\begin{tabular}{c|c|c|c|c}	
		\toprule
		\multicolumn{5}{c}{(a) SLCF performance} \\
		\midrule
		\multicolumn{1}{c|}{\multirow{2}[4]{*}{Conditions }} & \multicolumn{4}{c}{AP$^{B-box}$} \\
		\cmidrule{2-5}    \multicolumn{1}{c|}{} & AP$_{50}$  & AP$_{60}$  & AP$_{70}$  & AP  \\
		\midrule
		ResNet-101 backbone &  58 &  48.71 &  38.4 &  33.31 \\
		\multicolumn{1}{c|}{VGG-19 backbone} &  59.09  &  51.45 &  34.57 &  33.54 \\
		\midrule
		ResNet-101+SLCF &  58.99 &  52.26 &  40.45 &  33.91  \\
		VGG-19+SLCF &  61.64 &  53.69 &  39.05 &  33.91 \\
		
		\toprule
		\multicolumn{5}{c}{(b) SAG performance} \\
		\midrule
		\multicolumn{1}{c|}{\multirow{2}[4]{*}{Conditions }} & \multicolumn{4}{c}{AP$^{B-box}$} \\
		\cmidrule{2-5}    \multicolumn{1}{c|}{} & AP$_{50}$  & AP$_{60}$  & AP$_{70}$  & AP  \\
		\midrule
		ResNet-101 backbone &  58 &  48.71 &  38.4 &  33.31 \\
		\multicolumn{1}{c|}{VGG-19 backbone}  &  59.09  &  51.45 &  34.57 &  33.54 \\
		\midrule
		\multicolumn{1}{c|}{ResNet-101+SAG} &  58.08 &  50.22 &  39.37 &  33.88 \\
		\multicolumn{1}{c|}{VGG-19+SAG} &  59.33 &  51.53 &  36.29 &  34.26 \\
		\bottomrule
		
		\toprule
		\multicolumn{5}{c}{(c) SLCF+SAG performance} \\
		\midrule
		\multirow{2}[4]{*}{Conditions } & \multicolumn{4}{c}{AP$^{mask}$} \\
		\cmidrule{2-5}          & AP$_{50}$  & AP$_{60}$  & AP$_{70}$  & AP  \\
		\midrule
		\multicolumn{1}{c|}{ResNet-101 backbone} &  58 &  48.71 &  38.4 &  33.31 \\
		VGG-19 backbone  &  59.09  &  51.45 &  34.57 &  33.54 \\
		\midrule
		\multicolumn{1}{c|}{ResNet-101+SLCF+SAG} &  59.53 &  53.18 &  39.33 &  35.57 \\
		VGG-19+SLCF+SAG &  62.85 &  53.46 &  38.93 &  36.81 \\
		\bottomrule

		\toprule
		\multicolumn{5}{c}{(d) SLCF+SAG+SC performance} \\
		\midrule
		\multicolumn{1}{c|}{\multirow{2}[4]{*}{Conditions }} & \multicolumn{4}{c}{AP$^{B-box}$} \\
		\cmidrule{2-5}    \multicolumn{1}{c|}{} & AP$_{50}$  & AP$_{60}$  & AP$_{70}$  & AP  \\
		\midrule
		ResNet-101 baseline &  58 &  48.71 &  38.4 &  33.31 \\
		\multicolumn{1}{c|}{VGG-19 backbone} &  59.09  &  51.45 &  34.57 &  33.54 \\
		\midrule
		\multicolumn{1}{c|}{ResNet-101+SLCF+SAG+SC} &  60.01 &  53.87 &  39.28 &  35.67 \\
		\multicolumn{1}{c|}{VGG-19+SLCF+SAG+SC} & \textbf{ 62.97} & \textbf{ 56.46} & \textbf{42.01} & \textbf{37.62} \\
		\bottomrule
	\end{tabular}%
	\label{DETECTION}%
\end{table}%

\begin{table}[htbp]
	\centering
	\caption{Computation complexity comparison between the different combination of proposed modules}
	\begin{tabular}{c|c|c}
		\toprule
		\multicolumn{1}{c|}{Methods} & Parameters(M) & Flops(M) \\
		\midrule
		\multicolumn{1}{c|}{ResNet-50 backbone} & 26.54 & 53.04 \\
		ResNet-101 backbone & 31.34 & 62.59 \\
		\multicolumn{1}{c|}{VGG-19 backbone} & 25.3  & 50.6 \\
		\midrule
		ResNet-101+SLCF & 34.29 & 68.49 \\
		ResNet-101+SAG & 34.29 & 68.49 \\
		ResNet-101+SLCF+SAG & 34.36 & 68.62 \\
		VGG-19+SLCF & 28.24 & 56.5 \\
		VGG-19+SAG & 26.08 & 52.17 \\
		VGG-19+SLCF+SAG & 28.31 & 56.63 \\
		\midrule
		ResNet-101+SLCF+SAG+SC & \textbf{34.65} & \textbf{69.22} \\
		VGG-19+SLCF+SAG+SC & \textbf{29.88} & \textbf{59.77} \\
		\bottomrule
	\end{tabular}%
	\label{flops}%
\end{table}%

\textbf{Feature extraction network.} 
The performance of using different backbone (including VGG-19 \cite{simonyan2014very}, ResNet-50 \cite{he2016deep}, and ResNet-101) as feature extraction networks  is first evaluated. Both VGG-19 backbone and ResNet-101 backbone are used as the baselines in the experiments. ResNet has been proved to be more useful and efficient when comparing to VGG-19, but as shown in Table \ref{backbone}, VGG-19 performs better. With the fewest parameters, the VGG-19  backbone achieves segmentation performances of 49.33, 45.19,  35.21, and 30.97 at AP$_{50}$, AP$_{60}$, AP$_{70}$, and AP, respectively. As for detection performance, the VGG-19 backbone also reaches the best scores of 59.09, 51.45, 34.57, and 33.54 at AP$_{50}$, AP$_{60}$, AP$_{70}$, and AP, respectively. 
The main reason is that the ultrasound image has low resolution and less semantic information. When stacking many convolution layers,  the edges and shapes of targets might be discarded. 

\textbf{Spatial Local Contrast Feature (SLCF).}
This experiment mainly discuss the improvement performance of SLCF.
As shown in Table \ref{SEGMENTATION} (a) and \ref{DETECTION} (a), comparing to the baseline with VGG-19 backbone, SLCF increase the segmentation results by 5.2, 1.06, 0.94, and 2.24 on AP$_{50}$, AP$_{60}$, AP$_{70}$ and AP respectively. 
For detection results, SLCF improves the results by 4.73, 3.43, 1.34, and 4.23 on AP$_{50}$, AP$_{60}$, AP$_{70}$ and AP, respectively.  
It  indicates that  considering both local and spatial contrast is important and useful for segmentation and detection in ultrasound images.

\textbf{Self Attention Gate (SAG).}  
As shown in Table \ref{flops}, the increased parameters of adding SAG on VGG-19 backbone is only 0.78M. The results of using SAG module alone is shown in Table \ref{SEGMENTATION} \ref{DETECTION}  (b). 
Comparing the results in Table \ref{SEGMENTATION} (a) (b) and Table \ref{SEGMENTATION} (c), it can be found that SLCF and SAG can promote each other when used together,  and it is better than using SLCF or SAG alone. 

\textbf{Skip Concatenation with Transpose Convolution (SC).} 
To further demonstrate  the effectiveness of the synergy of SC with SLCF and SAG. In this experiment, the SC module is combined  with SLCF and SAG.  Because the transpose convolution instead of bilinear interpolation is adopted  to upsample the feature maps, the parameters are slightly increased. The results of computation complexity comparison in Table \ref{flops} show that for ResNet-101+SLCF+SAG+SC, SC only increases 0.36M parameters comparing to ResNet-101+SLCF+SAG. But comparing their segmentation results in Table \ref{SEGMENTATION} (c) and (d), SC greatly improves the performance by 3.96, 0.3, 5.51 and 1.98 on AP$_{50}$, AP$_{60}$, AP$_{70}$ and AP respectively, which indicates that  three modules can cooperate well with each other. As shown in Figure \ref{fig4}, combining all the proposed modules can greatly improve the performance on both segmentation and detection tasks. 

It is important to constrain the parameters of network and  while improving performance, so the comparison of parameters and FLOPs (Floating Point Operations) between the proposed method and baseline is evaluated.  
As shown  in Table \ref{flops}, the proposed  network  only increase a little computational complexity, but has greatly improved the performance on both segmentation and detection tasks. On the other hand, it also proves that the proposed SLCF and SAG are very lightweight and easy to use. 

\begin{figure*}[!t]
	\centerline{\includegraphics[width=0.8\textwidth]{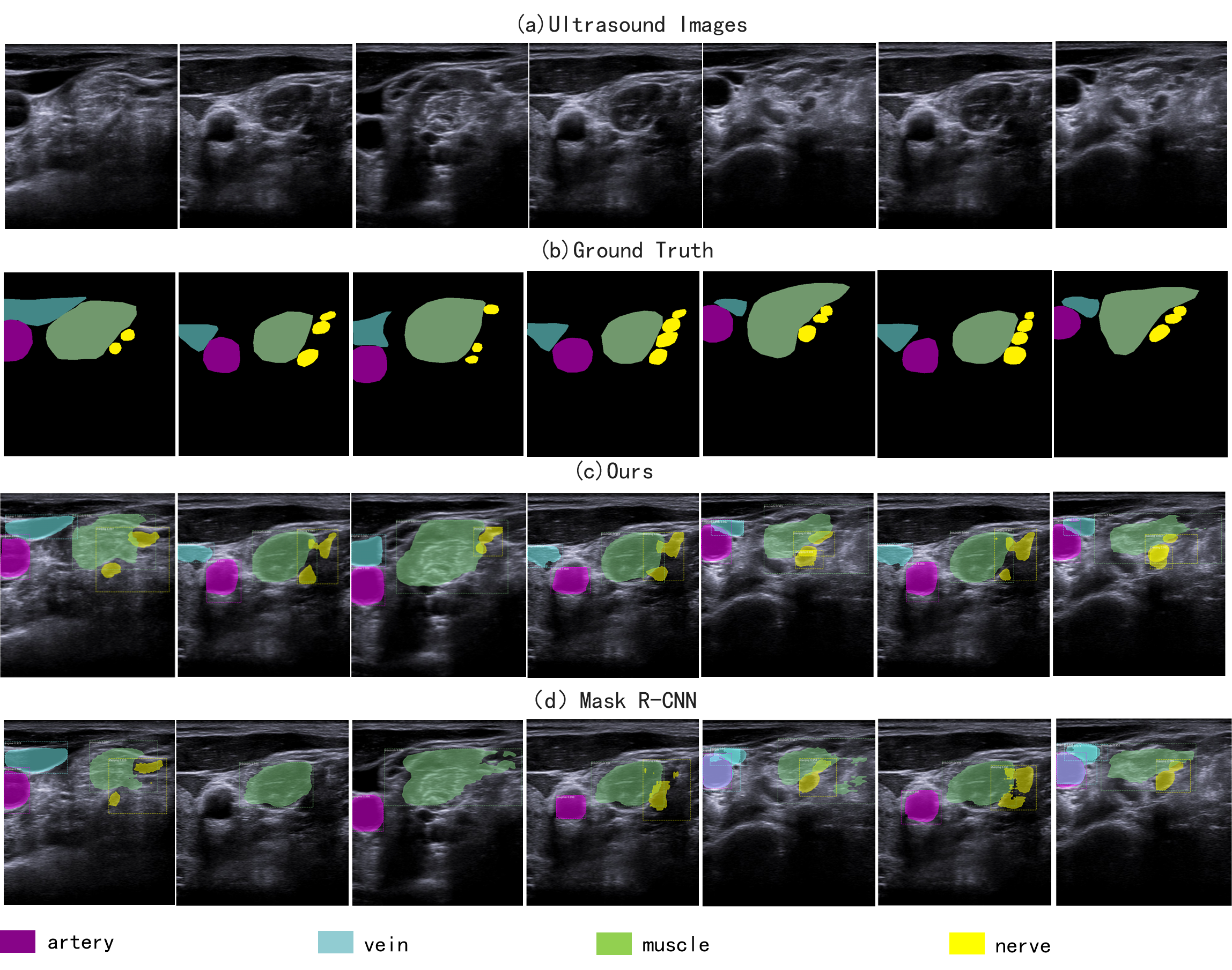}}
	\caption{
		Visualization of part segmentation results using different networks. (c) and (d) are the segmentation results of the proposed network and the Mask R-CNN.	
	}
	\label{results_pic}
\end{figure*}

\begin{figure}[!t]
	\centerline{\includegraphics[width=\columnwidth]{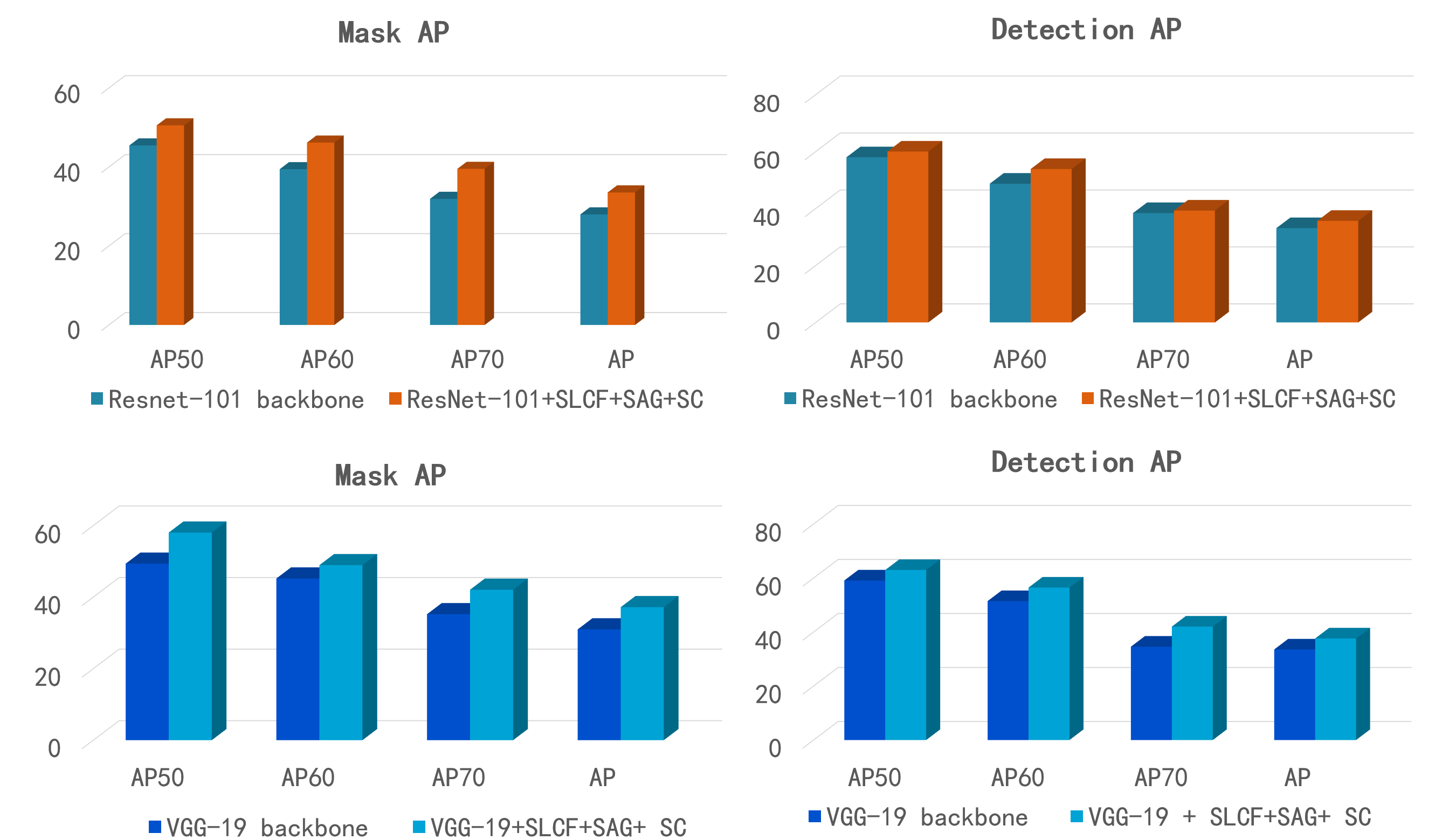}}
	\caption{
		Comparing the effect of the network with all proposed modules and without. 
	}
	\label{fig4}
\end{figure}

\begin{table}[htbp]
	\centering
	\caption{Performance comparison to different state-of-the-art methods on the test set of UBPD. The best results are
		highlighted in bold.}
	\begin{tabular}{c|c|c|c|c}
		\toprule
		\multicolumn{1}{c|}{\multirow{2}[4]{*}{Conditions }} & \multicolumn{4}{c}{AP$^{mask}$} \\
		\cmidrule{2-5}    \multicolumn{1}{c|}{} & AP$_{50}$  & AP$_{60}$  & AP$_{70}$  & AP  \\
		\midrule
		\multicolumn{1}{c|}{Mask R-CNN} & 43.96  & 39.45 & 30.41 & 27.89 \\
		\multicolumn{1}{c|}{YOLACT} & 57.15 & 45.95 & \textbf{42.81} & 34.02 \\
		\multicolumn{1}{c|}{OURS}  & \textbf{58.02} & \textbf{48.92} & 42.06 & \textbf{37.18} \\
		\midrule
		\midrule
		\multicolumn{1}{c|}{\multirow{2}[4]{*}{Conditions }} & \multicolumn{4}{c}{AP$^{B-box}$} \\
		\cmidrule{2-5}    \multicolumn{1}{c|}{} & AP$_{50}$  & AP$_{60}$  & AP$_{70}$  & AP  \\
		\midrule
		\multicolumn{1}{c|}{Mask R-CNN} & 58 & 48.71 & 38.4 & 33.11 \\
		\multicolumn{1}{c|}{YOLACT} & 62.01 & 51.58 & 41.28 & 30.39 \\
		\multicolumn{1}{c|}{OURS}  & \textbf{62.97} & \textbf{56.46} & \textbf{42.01} & \textbf{37.62} \\
		\bottomrule
	\end{tabular}%
	\label{others}%
\end{table}%

\subsection{Main Results}


Comparison experiments with Mask R-CNN and YOLACT have also been conducted. They are both state-of-the-art methods for image instance segmentation, and all methods are implemented according to the originally provided code.  Table \ref{others} presents the comparison results. 
As shown in Table \ref{others}, for detection task, our method achieve the best results at 62.97, 56.46, 42.01, and 37.62 on AP$_{50}$, AP$_{60}$, AP$_{70}$ and AP.  The proposed method achieves great results on segmentation task too, it exceeds other methods in segmentation task with the highest scores at 58.02, 48.92, and 37.18 on AP$_{50}$, AP$_{60}$, and AP. Comparing to the Mask R-CNN and YOLACT, the proposed method shows the consistently stability and availability for both segmentation and detection tasks.

The visualization of segmentation results on the test set of UBPD is shown in the Figure \ref{results_pic},  and each row has 7 sampled frames from  different  videos. The  Figure \ref{results_pic} (a) shows the raw ultrasound images blurred by noise interference, in which the nerves are too inconspicuous to identify. The segmentation results of the proposed network are shown in Figure \ref{results_pic} (c), which are very close to ground truth. This indicates that the proposed network can effectively detect and segment different tissues (nerves, arteries, veins, muscles) in ultrasound images, and generate the segmentation results with finer details and smoother edges. 
However,  comparing to the proposed network, the Mask R-CNN fails to effectively segment the nerve and other tissues. As shown in Figure \ref{results_pic} (d), there are instances with wrong segmentation categories and some are even not been identified. 
These comparison results prove the effectiveness of the proposed network. and the BPMSegNet can achieve a stable and better performance on Brachial Plexus instance segmentation tasks.

\section{Conclusion}
This paper addresses the problems in  brachial plexus ultrasound image segmentation to effectively assist anesthesiologist in PNB. 
According to the nerve identification process in clinical PNB, the identification of brachial plexus can be  converted into segmenting the nerve and its surrounding tissues.
But in the ultrasound images, these targets have different scales and reciprocal with each other, thus resulting in difficulties for ultrasound image segmentation. Therefore,  a novel method is proposed to segment multiple instances  in ultrasound image. 
Specifically,  the first ultrasound image dataset of brachial plexus (UBPD) and a novel network (BPMSegNet)   are proposed for the segmentation of brachial plexus and its surrounding anatomy. 
Quantitative experiments are carried out to verify the superiority of the proposed network on test set of UBPD,  and the network achieves state-of-the-art consistently  in both detection and segmentation tasks.

\section*{Acknowledgment}

This work was supported by the Natural Science Foundation of Guangdong Province (Grant No. 2018A030313354), the Neijiang Intelligent Showmanship Service Platform Project (No. 180589), the Sichuan Science-Technology Support Plan Program (No.2019YJ0636, No.2018GZ0236, No.18ZDYF2558), and the National Science Foundation of China - Guangdong Joint Foundation (No.U1401257).

\ifCLASSOPTIONcaptionsoff
  \newpage
\fi



%

\bibliography{reference}
\bibliographystyle{IEEEtrans}

\end{document}